\documentclass[10pt,conference,letterpaper]{IEEEtran}
\IEEEoverridecommandlockouts
\usepackage{cite}
\usepackage{amsmath,amssymb,amsfonts}
\usepackage{algorithmic}
\usepackage{graphicx}
\usepackage{textcomp}
\usepackage{subfigure}
\usepackage{float}
\usepackage{multirow}
\usepackage[table]{xcolor}
\usepackage{booktabs}
\usepackage{makecell}
\usepackage[ruled,linesnumbered]{algorithm2e}

\def\BibTeX{{\rm B\kern-.05em{\sc i\kern-.025em b}\kern-.08em
    T\kern-.1667em\lower.7ex\hbox{E}\kern-.125emX}}
\allowdisplaybreaks[4]

\begin{document}

\title{
AFLoRA: Adaptive LoRA for Resource-Efficient Federated Fine-Tuning of Large Language Models
}

\author{\IEEEauthorblockN{Yajie Zhou$^ {\dagger}$, Xiaoyi Pang$^ {\dagger}$, Zhibo Wang$^ {\dagger,}$\thanks{Correspondence to: yajiezhou@zju.edu.cn}}

}
\maketitle

\begin{abstract}
LoRA-based federated fine-tuning allows clients to fine-tune LoRA adapters locally and aggregates updates on the server, thereby enabling the efficient adaptation of global foundation models to distributed, domain-specific data.
However, existing methods assume identical and static LoRA adapters across clients, which not only limits the utilization of heterogeneous resources and adaptation to diverse data distributions, but also prevents dynamic adjustment to clients’ evolving local fine-tuning state.
They also struggle to achieve precise and lightweight aggregation of LoRA matrices, hindering learning performance in resource-constrained environments. 
To address these issues, we propose AFLoRA, an adaptive LoRA-based resource-efficient federated fine-tuning framework. 
For effectively adapting to Non-IID local data while minimizing resource consumption, AFLoRA introduces diagonal-based dynamic rank assignment, which dynamically adjusts local LoRA ranks based on client resources and local adaptation state guided by a learnable diagonal matrix. 
It further decouples LoRA matrices updates between clients and the server to reduce overhead, and develops a zero-padding-based rank-aware aggregation mechanism to accurately integrate heterogeneous updates through lossless matrix expansion and rank-aware weighting. 
These mechanisms enable AFLoRA to balance resource efficiency and fine-tuning performance effectively. 
Extensive experiments demonstrate that AFLoRA outperforms state-of-the-art methods in both LLM adaptation performance and fine-tuning efficiency in heterogeneous environments.

\end{abstract}

\section{Introduction}\label{sec:introduction}

Large Language Models (LLMs) have demonstrated strong general capabilities across a wide range of tasks but often underperform in domain-specific scenarios\cite{it1, it2, it3}.
Federated Learning (FL) \cite{federated,federated2,federated3} addresses this challenge by enabling multiple clients to collaboratively train a shared global model without exchanging raw data, thus facilitating LLM adaptation with distributed, private and domain-specific data.
To make fine-tuning such large-scale models feasible in resource-constrained edge environments, Parameter-Efficient Fine-Tuning (PEFT) methods \cite{adapter, bitfit, prefix, lora}, particularly Low-Rank Adaptation (LoRA) \cite{lora, lori, adalora, lorasb}, have become mainstream in federated fine-tuning scenarios. These methods freeze most LLM parameters and update only a small subset or lightweight adapters, dramatically reducing resource requirements and enabling efficient collaborative learning without data centralization, even on limited devices.


Despite their efficiency, most existing PEFT-based federated fine-tuning \cite{slora,fedadapter,fedbiot} methods assume that all clients adopt homogeneous fine-tuning modules for easy model update aggregation. 
This assumption is problematic for reasons as follows.
First, under system heterogeneity, homogeneous adaptation bottlenecks both communication and computation efficiency by constraining all clients to operate at the level of the most resource-limited device.
Second, under data heterogeneity, using identical fine-tuning modules across clients with Non-Independent and Identically Distributed (Non-IID) data can lead to suboptimal local adaptation, thereby hindering the generalization capability of the global model. 
In addition, we present a new perspective: As federated fine-tuning progresses, each client deepens its understanding of local data and the implicit knowledge. Consequently, the proper size of the fine-tuning module required to effectively capture local knowledge may also change dynamically over time. 
So maintaining a fixed module size for fine-tuning throughout the process could lead to inefficient use of client-side resources and potential waste.
These motivate an adaptive federated fine-tuning strategy applicable for real-world heterogeneous and resource-limited scenarios, which dynamically adjusts each client's fine-tuning module size according to their communication, computation, and data resources and local learning state to fully leverage their resources and learn from their local data.

Focusing on LoRA-based federated fine-tuning, beyond the problem of assuming homogeneous fine-tuning modules, another key issue is that classic averaging during global aggregation can introduce interference \cite{ffalora, flexlora, flora}.
Specifically, the LoRA parameter is the product of two low-dimensional parameter matrices, that is, $\mathbf{\Delta W}=\mathbf{B} \mathbf{A}$, where $\mathbf{W} \in \mathbb{R}^{m \times n}$, $\mathbf{A} \in \mathbb{R}^{r \times n}$, $\mathbf{B} \in \mathbb{R}^{m \times r}$, and $r$ is the rank of LoRA metrics. It represents the low-rank adjustment to the original weight matrix. 
Existing works usually average the LoRA matrices $\mathbf{A}$ and $\mathbf{B}$ separately in the aggregation process, 
yielding results inconsistent with averaging the complete matrix $\mathbf{\Delta W}$ (i.e., $\mathbf{B} \mathbf{A}$), which introduces drift between local and global updates and thus undermines the learning process of the global model. 
Recent works have designed new aggregation schemes for accurate LoRA aggregation, but most of them often introduce significant communication or computation overhead since they require the exchange of more parameters or extra calculation operations\cite{flexlora, flora}. 
To achieve efficient federated fine-tuning in resource-limited scenarios, there is an urgent need for new federated LoRA fine-tuning approaches that can accurately aggregate LoRA adapters of clients while conserving resources. 

In this paper, focusing on the above issues, our goal is to enable resource-efficient and accurate federated LoRA fine-tuning of LLMs in heterogeneous environments. 
To achieve this goal, we identify two core challenges: 
1) \textit{How to assign adaptive LoRA to clients for efficient LLM adaptation on local data while conserving client-side computation and communication resources?}
Due to heterogeneous local data and evolving learning states (i.e., how well the model has already adapted to the local data), the optimal LoRA rank size for each client is difficult to determine and varies over time. This challenge is further compounded by clients’ diverse and limited resources, making it hard to balance local adaptation needs with resource constraints.
2) \textit{How to achieve lightweight and accurate aggregation for heterogeneous LoRA adapters to enable efficient federated fine-tuning?}
Existing methods try to reduce interference in LoRA aggregation by changing aggregation strategies, but this often increases communication or computation overhead. Consequently, there is a challenging trade-off between aggregation accuracy and efficiency, especially in heterogeneous environments with varying LoRA ranks and data distributions.

To address the challenges, we propose AFLoRA, an adaptive LoRA-based resource-efficient federated fine-tuning framework, to achieve efficient and accurate LLM adaptation in heterogeneous environments while conserving client-side resources.
To this end, we integrate a learnable diagonal matrix $\mathbf{\Lambda} \in \mathbb{R}^{r \times r}$ into the LoRA framework, placed between the low-rank matrices $\mathbf{A}$ and $\mathbf{B}$. The LoRA weight update is redefined as $\mathbf{\Delta W}=\mathbf{B} \mathbf{\Lambda} \mathbf{A}$, where $\mathbf{\Lambda}$ introduces rank-wise scaling, enhancing the flexibility of the adaptation process.
In this framework, we first propose a Diagonal-Based Dynamic Rank Assignment, which utilizes $\mathbf{\Lambda}$ as a contribution evaluator for each dimension within the LoRA matrices and dynamically prunes individual less informative dimensions to adjust the effective LoRA rank based on local resources and adaptation state.
Second, we propose a Decoupled LoRA Fine-Tuning, where $\mathbf{B}$ and $\mathbf{\Lambda}$ are fine-tuned on clients and $\mathbf{A}$ is fine-tuned on the server, reducing client-side overhead and balancing generalization with local adaptation.
Third, we propose a Zero-Padding-Based Rank-aware Aggregation, which aligns heterogeneous LoRA metrics via padding and performs rank-aware weighted aggregation to accurately capture local knowledge. As the shared parts of $\mathbf{A}$ are identical, directly averaging the corresponding $\mathbf{B}$ updates enables exact aggregation.

Our main contributions are summarized as follows.

\begin{itemize}
  \item We propose AFLoRA, a lightweight and adaptive federated fine-tuning framework that reconfigures the LoRA architecture through decoupled adaptation and dynamic rank assignment. This framework enables efficient and accurate model adaptation across heterogeneous and resource-constrained clients.
  \item We propose a diagonal-based adaptive rank mechanism that dynamically adjusts the rank according to client resources and training status, thereby reducing redundancy. By decoupling adaptation, performing server-side tuning on matrix $\mathbf{A}$ and client-side tuning on matrix $\mathbf{B}$ and $\mathbf{\Lambda}$, we not only reduce client-side costs, but also effectively avoid interference between local and global updates.
  \item Extensive experiments demonstrate that AFLoRA outperforms state-of-the-art alternatives in model performance and client-side costs, achieving the same or even higher global model accuracy while reducing client-side communication and computation overhead by roughly half.
\end{itemize}



\section{Related Work}\label{sec:related}
In this section, we introduce the related work of FL with PEFT,  and LoRA-based federated fine-tuning in heterogeneous environments.

\subsection{Federated Learning with Parameter-Efficient Fine-Tuning}
PEFT refers to a family of techniques that enable the adaptation of large pre-trained models to downstream tasks by updating only a small subset of additional or re-parameterized components, rather than the entire model.
This makes PEFT particularly well-suited for FL, where clients typically have limited computation and communication resources.
Representative methods include FedAdapter\cite{fedadapter} using adapter tuning which inserts small trainable adapters into selected layers, FedIT\cite{fedit} using LoRA tuning which applies low-rank decomposition to weight matrices, pFedprompt\cite{pFedPrompt} using prompt tuning which prepends learnable prompt tokens to the input, or FedPrefix\cite{fedperfix} using prefix tuning which tunes prefix vectors in the transformer layers. 
LoRA is arguably the most popular method among PEFT approaches, requiring the adjustment of less than 1\% of model parameters while achieving performance comparable to full fine-tuning across a wide range of tasks. 
Building on the success of LoRA, various works\cite{openfedllm, fedsb, fedsalora, fedlfc} have integrated it into FL. 
Variants such as SLoRA\cite{slora}, FedSA-LoRA\cite{fedsalora}, and FedLFC\cite{fedlfc} improve LoRA-based FL by enhancing initialization, selectively aggregating parameters, or enabling multi-task adaptation through clustering.

\subsection{LoRA-based Federated Fine-Tuning in Heterogeneous Environments}
Heterogeneity in resources such as computation, communication, and data poses major challenges in FL. 
Existing methods have also begun to support heterogeneous or adaptive LoRA configurations to address resource heterogeneity in FL, but limitations remain. Attempts like FlexLoRA \cite{flexlora} and FLoRA \cite{flora} select LoRA ranks based on local resources but fix them throughout training, ignoring changes in data and learning dynamics. HETLoRA \cite{hetlora} uses regularization-based pruning but adds hyperparameter complexity, while PF2LoRA \cite{pf2lora} and AutoRank \cite{autorank} lack adaptivity or are overly complex. These approaches struggle to balance efficiency, adaptability, and performance under heterogeneous and constrained environments.
Meanwhile, LoRA-based federated fine-tuning methods often suffer from aggregation interference, i.e. discrepancies between averaged and local LoRA weights, which hinders global performance. Solutions like FlexLoRA\cite{flexlora}’s SVD reconstruction, FFA-LoRA\cite{ffalora}’s partial freezing, or FLoRA\cite{flora}’s matrix stacking introduce extra computation, limit capacity, or increase communication costs. Thus, accurately and efficiently aggregating heterogeneous LoRA adapters without sacrificing performance or scalability remains an open challenge.

\section{Preliminary}\label{sec:pre}
In this section, we first introduce LoRA-based federated fine-tuning, and then provide a brief overview of the relevant concepts of Singular Value Decompositionto (SVD) better introduce our proposed framework.

\subsection{LoRA-based Federated Fine-Tuning}
LoRA is a PEFT method that reduces communication and computation overhead by freezing the pre-trained backbone $\mathbf{W}^t$ and adding trainable adapters $\mathbf{A}^t$ and $\mathbf{B}^t$ to selected layers, computing the weight update $\mathbf{\Delta W}^t$ using low rank decomposition at global round $t$. $\mathbf{W}^t, \mathbf{\Delta W}^t \in \mathbb{R}^{m \times n}$, and $\mathbf{\Delta W}^t$ is parameterized by two low-rank matrices: $\mathbf{A}^t \in \mathbb{R}^{r \times n}$ and $\mathbf{B}^t \in \mathbb{R}^{m \times r}$. The $\mathbf{W}^t$ is updated as:

\begin{equation}\label{eq:lora}
\mathbf{W}^{t+1} = \mathbf{W}^{t} + \mathbf{\Delta W}^{t} = \mathbf{W}^{t} + \mathbf{B}^{t} \mathbf{A}^{t}.
\end{equation}
Since $r$ represents the rank of the low-rank matrices, and rank $r \ll \min(m,n)$, this decomposition drastically reduces the number of trainable parameters, making LoRA particularly suitable for resource-constrained environments.

Classic LoRA-based federated fine-tuning algorithm combines the FedAvg\cite{fedavg} framework with LoRA\cite{lora}. 
Let $\mathcal{G}^t$ donates the set of the participating clients at global round $t$.
Each client $k \in \mathcal{G}^t$ fine-tunes only the LoRA modules during local training, that is, the low-rank matrices $\mathbf{A}_k^t$ and $\mathbf{B}_k^t$ while keeping the pre-trained backbone model frozen. After a number of local training steps, client $k$ uploads the updated $\mathbf{A}_k^t$ and $\mathbf{B}_k^t$ to the server. The server then performs a weighted aggregation of the parameters using FedAvg, where each client’s update is scaled by its corresponding weight $p_k$ representing its data proportion:
\begin{equation}\label{eq:avg}
\mathbf{A^t} = \sum_{k \in \mathcal{G}^t} p_k \mathbf{A}_k^t, \quad
\mathbf{B^t} = \sum_{k \in \mathcal{G}^t} p_k \mathbf{B}_k^t.
\end{equation}
Then the server updates $\mathbf{W}$ based on Eq. \eqref{eq:lora} and broadcasts the aggregated $\mathbf{A}^t$ and $\mathbf{B}^t$ back to clients. For each client $k$, it updates its local foundation model as:
\begin{equation}\label{eq:avg}
\mathbf{W}_k^{t+1} = \mathbf{W}^{t} + \mathbf{B}^{t} \mathbf{A}^{t}.
\end{equation}

\subsection{Singular Value Decomposition}
SVD is a fundamental matrix factorization technique widely used in linear algebra and machine learning. For any real matrix $\mathbf{W} \in \mathbb{R}^{m \times n}$, the SVD is formulated as:
\[
\mathbf{W} = \mathbf{U \Sigma V^\top},
\]
where $\mathbf{U} \in \mathbb{R}^{m \times r}$ and $\mathbf{V} \in \mathbb{R}^{n \times r}$ are column-orthogonal matrices, and $r$ denotes the rank of $\mathbf{W}$. The diagonal matrix $\mathbf{\Sigma} \in \mathbb{R}^{r \times r}$ contains the singular values $\sigma_1 \geq \sigma_2 \geq \dots \geq \sigma_r > 0$ on its diagonal. Each singular value $\sigma_i$ represents the information content, i.e. the amount of information or energy that $\mathbf{W}$ retains along the corresponding latent dimension.

\section{Problem Definition}\label{sec:pro}
\begin{figure}[!t]
\centering
\includegraphics[width=0.6\columnwidth, height=1.5in, keepaspectratio=false]{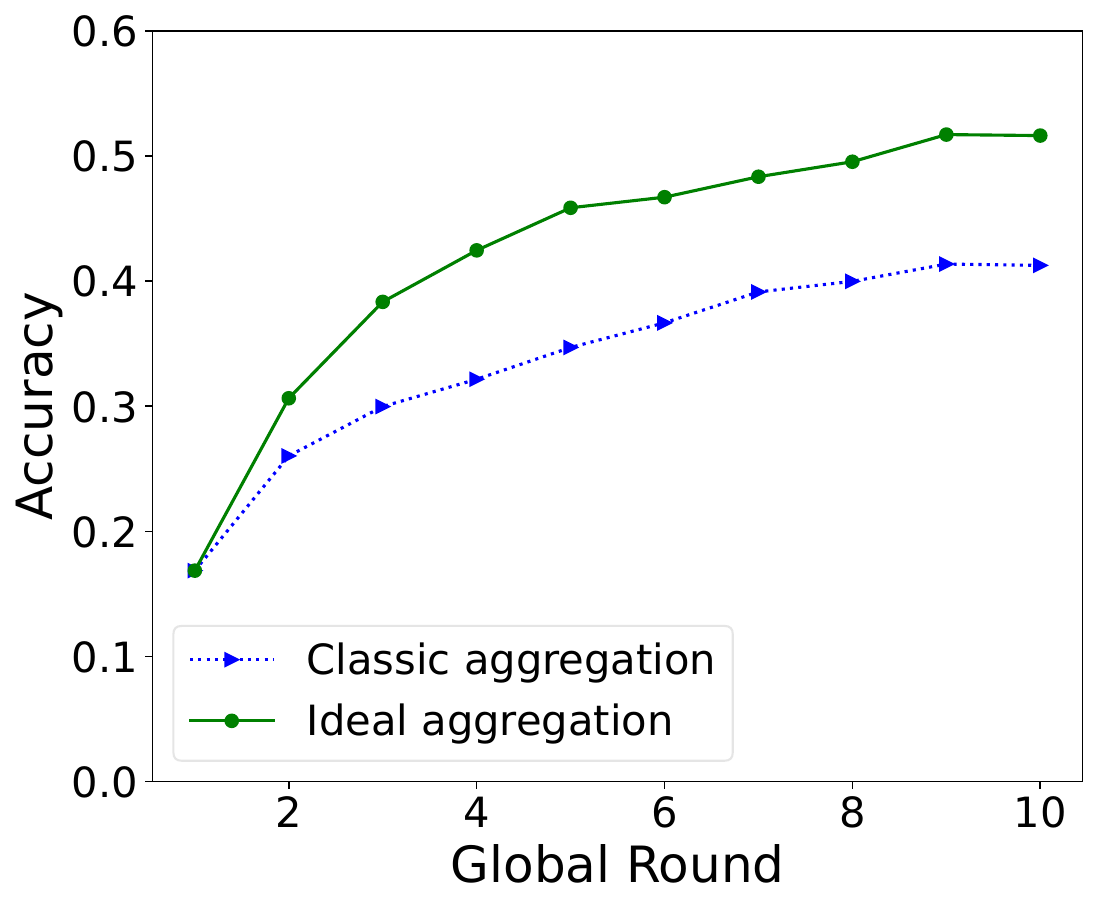}
\vspace{-3mm}
\caption{Performance comparison under different aggregation methods.}
\label{fig:motivation}
\vspace{-5mm}
\end{figure}
In this section, we first define our system model and analyze the motivation. Then we formally present the problem to be solved in this paper.

\subsection{System Model}
In this paper, we investigate a federated fine-tuning system designed for heterogeneous and resource-limited scenarios, consisting of a central server and $K$ clients with varying and constrained computation and communication capabilities. 
Each client $k \in U=\{1, 2, \ldots, K\}$ possesses a local dataset $D_k= \{(x_{k,i}, y_{k,i})\}_{i=1}^{|D_k|}$, where $|D_k|$ denotes the size of client $k$’s dataset and the data distribution across clients is Non-IID. The central server maintains a small-scale public dataset $D_{{public}}$ that is entirely disjoint from all client datasets, i.e., $D_{{public}} \cap D_k = \emptyset$ for all $k \in \{1, 2, \ldots, K\}$.
During training, each client fine-tunes LoRA adapters locally and uploads model updates to the server, which then aggregates these updates to update the global model. The federated fine-tuning objective is formulated as:
\begin{equation}\label{eq:obj}
\min_{\mathbf{W}+\mathbf{B} \mathbf{A}} F(\mathbf{W}+\mathbf{B} \mathbf{A}) = \sum_{k=1}^K p_k F_k(\mathbf{W}+\mathbf{B} \mathbf{A}),
\end{equation}
where $p_k=\frac{|D_k|}{|D|}$, $\mathbf{W}$, $\mathbf{B}$ and $\mathbf{A}$ are the global model and two global adapters, $F_k(\mathbf{W}+\mathbf{B} \mathbf{A})$ is the local loss of client $k$ and $|D| = \sum_{i=1}^K |D_i|$ is the total number of data samples across all clients.


\begin{figure*}[ht]
\centering
\label{fig:overview}
\includegraphics[width=1.65\columnwidth]{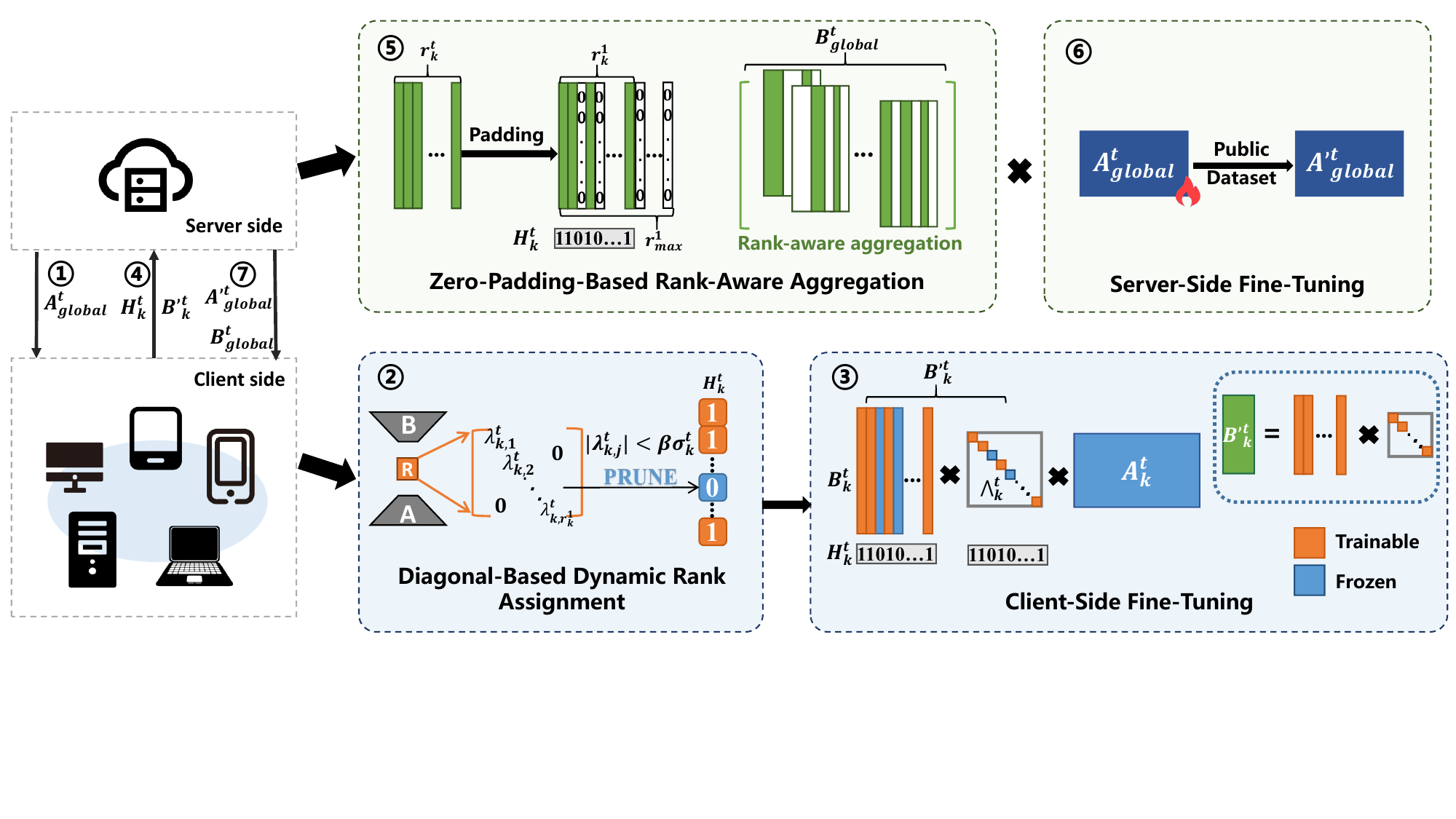}
\vspace{-2.5mm}
\caption{The overview of AFLoRA}
\label{fig:overview}
\vspace{-5mm}
\end{figure*}

\subsection{Motivation}
According to the principle of FL, the ideal approach is to aggregate all clients' updates directly, i.e., ideal aggregation, which can be mathematically expressed as $\mathbf{\Delta W}^t=\sum_{k \in \mathcal{C}^t} p_k (\mathbf{B}_k^t \cdot \mathbf{A}_k^t)$.
However, this is rarely adopted in practice because it incurs significant computation and communication overhead: computing the matrix product for each client requires as many multiplications as the number of clients, and transmitting the full matrix $\mathbf{\Delta W}^t$ (with the same size as the original model weight $\mathbf{W}$) is much more costly than transmitting the low-rank adapters $\mathbf{A}^t$ and $\mathbf{B}^t$.
Although classic aggregation strategy of LoRA-based federated fine-tuning can solve this problem, it introduces an inherent interference problem.
When we aggregate these components separately across all $K$ clients and then compute their product, we encounter a mathematical discrepancy:
\begin{equation}
\mathbf{\Delta W}^t=\underbrace{ 
\left( \sum_{k \in \mathcal{G}^t} p_k \mathbf{B}_k^t \right) \left( \sum_{k \in \mathcal{G}^t} p_k \mathbf{A}_k^t \right)
}_{\text{Classic aggregation}}
\neq
\underbrace{
 \sum_{k \in \mathcal{G}^t} p_k (\mathbf{B}_k^t \mathbf{A}_k^t)
}_{\text{Ideal aggregation}}.
\end{equation}
As a result, the aggregated update significantly deviates from the optimal direction that would have been obtained by directly averaging $\mathbf{B}_k^t \mathbf{A}_k^t$ across clients.
This bias between the global aggregation results and local updates leads to significant parameter drift. As a result, model convergence is severely compromised, and the global model may ultimately fail to capture the true local data distribution.
The severity of this interference effect is clearly demonstrated in our empirical analysis in Figure~\ref{fig:motivation}, where the performance gap between classic aggregation and ideal aggregation becomes increasingly pronounced as training progresses. 

\subsection{Problem Formulation}
Suppose there is a LoRA-based federated fine-tuning system consisting of a server and $K$ heterogeneous and resource-constrained clients.
During each global round $t$, each participating client $k \in \mathcal{G}^t$ fine-tunes its local LoRA adapters and uploads the updates to the server. 
Due to varying resources, each client fine-tunes and uploads LoRA adapter matrices with customized ranks, resulting in different dimensions across clients.
The server then aggregates all heterogeneous updates to get new global LoRA adapters, which are then used to update the foundation model and broadcast to clients for the next global round.
In this context, the problem is subject to several key constraints. First, each client can only use a limited LoRA rank due to computation and communication limitations. Second, clients differ in both data distribution and available resources, resulting in heterogeneous LoRA adapters.
In summary, the problem to be solved in this paper is:
How to effectively achieve accurate federated fine-tuning across heterogeneous clients while conserving local computation and communication resources, such that the global model achieves minimal loss on downstream tasks.

\section{AFLoRA Framework}\label{sec:mechanism}

We develop AFLoRA, an adaptive and lightweight federated fine-tuning framework built upon LoRA, aiming to achieve efficient LLM adaptation under constrained and heterogeneous resources. In this section, we give an overview of AFLoRA and then introduce the proposed mechanisms in detail.

\subsection{Overview of AFLoRA}
As shown in Fig. \ref{fig:overview}, we propose a novel LoRA-based federated fine-tuning framework AFLoRA, which consists of three mechanisms.
First, we insert a trainable diagonal matrix between the $\mathbf{A}$ and $\mathbf{B}$ adapters of LoRA to measure the the information content across different dimensions.
By adaptively pruning low-information dimensions, our method enables efficient adaptation to heterogeneous resources and evolving learning conditions.
Second, we adopt a two-stage, decoupled LoRA update strategy: the $\mathbf{B}$ and $\mathbf{\Lambda}$ matrix is fine-tuned locally, while the $\mathbf{A}$ matrix is updated on the server, which balances generalization and local personalization while reducing client-side computation and communication overhead.
Third, to achieve accurate aggregation, we first pad heterogeneous $\mathbf{B}$ matrices to a uniform shape and then perform rank-aware weighted aggregation. As the shared parts of $\mathbf{A}$ are identical, the corresponding $\mathbf{B}$ updates can be directly averaged, making the global update equivalent to local updates.

The workflow begins with the server initializing and distributing the global parameter $A^t_{{global}}$ to all participating clients $\mathcal{G}^t$. 
Each client $k \in \mathcal{G}^t$ then performs diagonal-based dynamic rank assignment to determine its mask vector $H_k^t$. 
Leveraging this mask, it conducts local fine-tuning, updating the trainable local adapters, and subsequently uploads them to the server. 
The server applies zero-padding to local updates according to the corresponding mask, followed by rank-aware aggregation to obtain the global parameter $B^t_{{global}}$.
Finally, the server further fine-tunes $A^t_{{global}}$ using a public dataset before broadcasting updated parameters $\mathbf{A'}^t_{{global}}$ with $B^t_{{global}}$ to clients to update local foundation model for the next communication round.

\subsection{Diagonal-Based Dynamic Rank Assignment}\label{sec:Adaptive Rank Assignment}

To enable resource-efficient federated fine-tuning, it is essential to dynamically adaptively assign LoRA ranks to individual clients based on their computation, communication, data resources, and learning status.
Our approach is guided by two principles:
First, the computation and communication overhead required by the assigned local LoRA should not exceed local resource constraints.
Second, considering that different learning needs impose varying requirements on the parameter scale of LoRAs, we need to dynamically adjust the LoRA rank according to the local learning state, so as to achieve efficient utilization of local resources. Based on these principles, we first determine the maximum allowable rank for each client according to its resource constraints and assign this value as their initial rank, i.e., $R^1=\{r^1_1, \ldots, r^1_K\}$.
And then we measure the information content of local adapters by a trainable diagonal matrix and adjust its rank dynamically.

Inspired by SVD, the core idea is that we introduce a learnable diagonal matrix $\mathbf{\Lambda}^t_k=\mathrm{diag}(\lambda^t_{k,1}, \ldots, \lambda^t_{k,r_k^1})$ between matrices $\mathbf{A}^t_k \in \mathbb{R}^{r^1_k \times n}$ and $\mathbf{B}^t_k \in \mathbb{R}^{m \times r^1_k}$ for each client $k$ in round $t$.
Each diagonal element $\lambda^t_{k,j}$ quantifies the information content of the $j$-th low-rank dimension, with only these diagonal elements being learnable.
In each global round, the diagonal elements of $\mathbf{\Lambda}^t_k$ are updated to reflect both the local data distribution and the current learning state. 
The information content of each element reflects its contribution to local updates, dynamically indicating which low-rank dimensions are active or redundant as learning progresses with different data distribution.
Based on the values of $\mathbf{\Lambda}^t_k$, dimensions with lower information content are progressively pruned, resulting in an adaptive adjustment of the LoRA rank for each client.

\textbf{Information measurement: } 
First, as in SVD, to ensure that the diagonal elements $\lambda^t_{k,j}$ of $\mathbf{\Lambda}^t_k$ can directly reflect the information content of each dimension without being affected by the scale of $\mathbf{A}^t_k$ and $\mathbf{B}^t_k$, we apply the normalization constraints $\| \mathbf{a}^t_{k,j} \|_2 = C$ and $\| \mathbf{b}^t_{k,j} \|_2 = 1$, where $\mathbf{a}^t_{k,j} \in \mathbb{R}^n$ denotes the $j$-th row of $\mathbf{A}^t_k$, $\mathbf{b}^t_{k,j} \in \mathbb{R}^m$ denotes the $j$-th column of $\mathbf{B}^t_k$, and $\lambda^t_{k,j}$ is the $j$-th diagonal element of $\mathbf{\Lambda}^t_k$. As a result, local LoRA updates can be formulated as
\begin{equation}
\Delta \mathbf{W}^t_k 
= \mathbf{B}^t_k \mathbf{\Lambda}^t_k \mathbf{A}^t_k 
= \sum_{j=1}^{r} \mathbf{b}^t_{k,j} \lambda^t_{k,j} \mathbf{a}^{t\top}_{k,j}.
\end{equation}
The contribution of each dimension $j$ to the local updates $\Delta \mathbf{W}^t_k$ can differ significantly. To quantify this, we define the concept of \textit{dimensional information content}, measured by the Frobenius norm.
We define the information content of the $j$-th dimension in the low-rank adaptation as the energy contributed by its rank-1 component, $\mathbf{b}^t_{k,j} \lambda^t_{k,j} \mathbf{a}^{t\top}_{k,j}$, to the general update of the parameters $\Delta \mathbf{W}^t_k$. Thus, the contribution of each component to the overall update can be quantified as
As a result, the contribution of each component to the overall update can be measured as
\begin{equation}
\begin{split}
\mathrm{I}^t_{k,j} &:= \| \lambda^t_{k,j} \mathbf{b}^t_{k,j} \mathbf{a}^{t\top}_{k,j} \|_F^2\\ &= \| \lambda^t_{k,j}\|^2 \cdot \| \mathbf{b}^t_{k,j} \|_2^2 \cdot \| \mathbf{a}^t_{k,j} \|_2^2 = C^2 \cdot \|\lambda^t_{k,j}\|^2.
\end{split}
\end{equation}
Thus, $\|\lambda^t_{k,j}\|^2$ can be interpreted as a soft surrogate for singular values in SVD, serving as a proxy for the contribution of each dimension to the update.

\textbf{Dynamic rank assignment: }
At the first round of federated fine-tuning, we determine the maximum rank for each client based on their computation and communication capacities, and for each client $k$, we initialize a mask vector $H_k = [1, 1, \ldots, 1]^\top \in \mathbb{R}^{r_k^1}$.
Subsequently, in each training round, we adjust each client's local LoRA rank by discarding low-contribute dimensions based on information content measured by the trainable diagonal matrix. 

Specifically, for each local LoRA adapter, we prune dimensions whose \textit{dimensional information content} is noticeably smaller than the majority and can be seen as outliers. Inspired by the $\beta$-$\sigma$ rule in statistics, we identify outliers based on the standard deviation.
At global round $t$, given the updated $\mathbf{\Lambda'}_k^t = \left\{ \|\lambda_{k,j}^t\|^2 \right\}_{j \,\mathrm{s.t.}\, h_j^{k, t} = 1}$, each participating client $k$ computes the standard deviation of $\mathbf{\Lambda'}_k^t$.
To determine which dimensions are retained for the next round, we update each $h_j^{k, t+1} \in H^{k, t+1}$ as follows:
\begin{equation}
h_j^{k, t+1} = h_j^{k, t} \cdot \mathbb{I}\left(\|\lambda_{k,j}^t\|^2 \geq \beta \sigma_k^t\right),
\end{equation}
where $\beta$ is a hyperparameter controlling the pruning threshold and $\mathbb{I}$ is the indicator function.
The updated rank is then given by the number of remaining dimensions:
\begin{equation}
    r_k^{t+1} = \sum_{j=1}^{r_k^1} H_j^k.
\end{equation}

This principle allows us to adaptively control the sparsity of the model, ensuring that only informative dimensions are retained. As a result, the rank of each client is dynamically adjusted during local fine-tuning, according to its current learning state and the distribution of information across dimensions.



\subsection{Decoupled LoRA Fine-Tuning}\label{sec:LoRA A Frozen Tuning}
Prior studies\cite{lori, fedsalora} have shown that LoRA fine-tuning exhibits parameter redundancy, where $\mathbf{A}$ is primarily responsible for capturing general knowledge and $\mathbf{B}$ focuses on learning client-specific information. 
To improve efficiency and reduce local communication and computation overhead, we propose to decouple the training of $\mathbf{A}$ and $\mathbf{B}$ in LoRA. Specifically, in AFLoRA, $\mathbf{B}$ and $\mathbf{\Lambda}$ are fine-tuned on the client side, while $\mathbf{A}$ is updated on the server side with the assistance of a public dataset and shared among all clients. 
Compared to conventional LoRA-based federated fine-tuning, this method reduces the costs of client-side communication and computation by almost half since the client no longer needs to update and upload the matrix $\mathbf{A}$. Besides, since $r \ll \min(m, n)$, the introduced diagonal matrix $\mathbf{\Lambda}$ is much smaller than $\mathbf{A}$.

\textbf{Client-Side Fine-Tuning: }
At the beginning of each training round $t$, each client $k$ truncates the latest global matrix $\mathbf{A}_{{global}}^t \in \mathbb{R}^{r_{{max}}^1 \times n}$ to its first $r_k^1$ rows, obtaining $\mathbf{A}k^t \in \mathbb{R}^{r_k^1 \times n}$, which is then kept fixed during local fine-tuning. The global matrix $\mathbf{A}_{{global}}^t$ is initialized by the server, with each row $\mathbf{a}_j$ normalized to satisfy the constraint $\|\mathbf{a}_j\|_2 = C$.
Then, each client $k$ initializes both $\mathbf{B}_k^t$ and $\mathbf{\Lambda}_k^t$ with zeros. According to the mask vector $H_k^t$, only the columns of $\mathbf{B}_k^t$ corresponding to $h_{k,j}^t = 1$ are selected for fine-tuning, while the remaining columns are kept frozen. Subsequently, $\mathbf{B}_k^t$ and $\mathbf{\Lambda}_k^t$ are updated using the local dataset $D_k$.
As we mentioned above, to ensure that the learned scaling factors $\lambda_j^t$ in $\mathbf{\Lambda}_k^t$ accurately reflect the significance of each adaptation direction $j$, we impose a normalization constraint on the columns of $\mathbf{B}_k^t$.
Specifically, the $\ell_2$ norm of each column vector $\mathbf{b}_j^t$ in $\mathbf{B}_k^t$ is regularized to remain close to $1$. Then, to update both $\mathbf{B}_k^t$ and $\mathbf{\Lambda}_k^t$, the objective for local fine-tuning on client $k$ can be formulated as:
\begin{equation}\label{finetuning loss}
\begin{split}
F_k(\mathbf{B}_k^t, \boldsymbol{\Lambda}_k^t) &= \frac{1}{|D_k|} \sum_{(x_{k,i}, y_{k,i}) \in D_k} F_{\text{CE}}(x_{k,i}, y_{k,i}) \\
&+ \gamma \cdot\sum_{j=1}^{r} \left( \| \mathbf{b}_{k,j}^t \|_2^2 - 1 \right)^2,
\end{split}
\end{equation}
$F_{\mathrm{CE}}(x_{k,i}, y_{k,i})$ is the cross-entropy loss for sample $(x_{k,i}, y_{k,i})$, defined as $F_{\mathrm{CE}}(x_{k,i}, y_{k,i}) = -\log \left( \frac{\exp(z_{k,i}^{(y_{k,i})})}{\sum_{l=1}^{|L|} \exp(z_{k,i}^{(l)})} \right)$, where $z_{k,i}^{(l)}$ denotes the model logit for class $l$ and $|L|$ is the number of classes.
and $\gamma$ is a hyperparameter controlling the strength of the normalization term.

Upon completing local training, each client $k$ obtains $\mathbf{B'}_k^t \in R^{m \times r_k^t}$ by multiplying the fine-tuned columns of $\mathbf{B}_k^t$, that is, those corresponding to $h_{k,j}^t = 1$, with the respective entries in $\mathbf{\Lambda}_k^t$.
The client then uploads only $\mathbf{B'}_k^t$ and $\mathbf{H}_k^t$ to the server. 
The inclusion of the diagonal matrix enriches the expressiveness of the update, thereby providing a more nuanced representation for server-side aggregation while preserving client adaptation flexibility.

\textbf{Server-Side Fine-Tuning: }
Given the typical Non-IID nature of client data distributions in federated settings, aggregation of only client-side updates risks overfitting the global foundation model to dominant or skewed distributions. 
To address this, we leverage a small, publicly available dataset to update the $\mathbf{A}$ matrix at the server side, trying to enhance the generalization ability of the global model.

At each round $t$, before updating the $\mathbf{A}^t_{global}$, the server aggregates $\{\mathbf{B'}_k^t\}_{k \in \mathcal{G}^t}$ to get $\mathbf{B}^t_{global}$ and freezes it during the fine-tuing process. Then the server updates $\mathbf{A}^{t}_{global}$ to get $\mathbf{A}^t_{FT}$ using the public training dataset based on the classic cross-entropy loss.
To strike a balance between preserving prior knowledge and incorporating new information, we fuse the original and fine-tuned $\mathbf{A}$ matrices to get the final $\mathbf{A'}^t_{global}$:
\begin{equation}
\mathbf{A'}_{global}^t = \alpha \cdot \mathbf{A}_{global}^{t} + (1 - \alpha) \cdot \mathbf{A}_{FT}^t, 
\end{equation}
where $\alpha \in [0,1]$ is a hyperparameter controlling the fusion ratio. 

Such a server-side fine-tuning of the $\mathbf{A}$ matrix offers two primary advantages: first, it preserves client-specific features by maintaining the personalized information encoded in $\mathbf{B}$, thus preventing the overwriting of local adaptations. Second, by leveraging a small public dataset to refine $\mathbf{A}$, this approach enhances the generalization ability of global model, effectively mitigating the challenges posed by data heterogeneity, reducing the risk of overfitting to specific client distributions, and alleviating catastrophic forgetting.

\subsection{Zero-Padding-Based Rank-Aware Aggregation}\label{sec:Heterogeneous LoRA Aggregation}

To aggregate heterogeneous local updates and better capture local knowledge, we propose the following mechanism: we first apply zero-padding to make all LoRA modules homogenous in rank, and then incorporate both the quantity of local data and the information content reflected by the rank of each client's update to achieve more precise aggregation.

\textbf{Zero-padding-based harmonization of LoRA: }
For $t$-th round of aggregation, we determine the maximum rank $r_{max}^t$ among all participating clients in $\mathcal{G}^t$.
Each client’s update $\mathbf{B'}_k^t$, originally of shape $m \times r_k^t$, is first expanded to $m \times r_k^1$ according to $H_k^t$: for each $j$ where $h_{k,j}^t = 0$, the $j$-th column in the expanded matrix is filled with zeros. Then, additional zero columns are appended to the right until the matrix reaches the shape $m \times r_{{max}}^1$, resulting in the padded matrix $\mathbf{B'}_k^{t, r_{{max}}^1}$.
This simple yet effective alignment ensures that all matrices are of equal dimensionality, enabling direct element-wise aggregation. Importantly, zero-padding operation does not introduce additional information, thereby retaining the integrity of each client’s learned features while allowing for integration across heterogeneous LoRA.
After padding, $\mathbf{B'}_k^{t, r_{{max}}^1} \times \mathbf{A}_{{global}}^t$ yields exactly the locally fine-tuned result.

\textbf{Rank-aware aggregation: }
After harmonization, we aggregate the padded LoRA matrices by assigning each client a weight based on both its local data volume and LoRA rank. The rationale is that a higher rank typically endows a client model with greater expressive capacity, potentially encoding more diverse or informative updates.
To balance this expressiveness with fairness, we propose a composite weighting scheme to arrange $p^t_k$ for client $k$ at round $t$:
\begin{equation}
p^t_k =  \frac{\log(1 + r_k^t)}{\sum_{k \in \mathcal{G}^t} [\log(1 + r_k^t)]} \cdot \frac{|D_k|}{\sum_{k \in \mathcal{C}^t} |D_k|}.
\end{equation}
Then we aggregate all updates as:
\begin{equation}
\mathbf{B}_{global}^t = \sum_{k \in \mathcal{G}^t} p^t_k \cdot \mathbf{B'}_k^{t,r^1_{max}}.
\end{equation}
This rank-aware aggregation scheme preserves the contribution of clients with higher-rank adapters and rich local data, avoiding the domination of updates by clients with larger adapters or datasets. 
Then the global update of the LoRA adapter can be computed by:
\begin{align*}
\mathbf{\Delta W}^t &= \mathbf{B}_{global}^t \times \mathbf{A}_{global}^t \\
&= \sum_{k=1}^{K} p^t_k \mathbf{B'}_k^{t,r^1_{max}} \times \mathbf{A}_{global}^t \\
&= \sum_{k=1}^{K} p^t_k \mathbf{B'}_k^{t,r^1_{k}} \times \mathbf{A}_{global}^t[{:}r^1_{k}, :] + \mathbf{0} \\
&= \sum_{k=1}^{K} p^t_k 
\mathbf{B'}_k^t \mathbf{A}_{k}^t = \sum_{k=1}^{K} p^t_k 
\mathbf{B}_k^t \mathbf{\Lambda}_k^t \mathbf{A}_{k}^t,
\end{align*}
where $ \mathbf{A}_{global}^t[{:}r^1_{k}, :]  $ is the sub-matrix consisting of the first \( r^1_{k} \) rows of \( \mathbf{A}^t_{global} \). Therefore, the aggregated result matches the client updates exactly, achieving ideal aggregation.

The server then broadcasts the updated $\mathbf{B}_{global}^t$ along with $\mathbf{A}'^t_{global}$ to all clients. 
Each client $k$ updates its local foundation model using the product $\mathbf{\Delta W}_k^t = \mathbf{B}_{global}^t \mathbf{A'}_{global}^t$ as the basis for subsequent local adaptation.

\section{Performance Evaluation}\label{sec:evaluation}
\begin{figure*}[!t]
\centering
\label{fig:ablation}
\subfigure[GPT-2]{\label{fig:ablation_gpt}
\includegraphics[width=0.625\columnwidth]{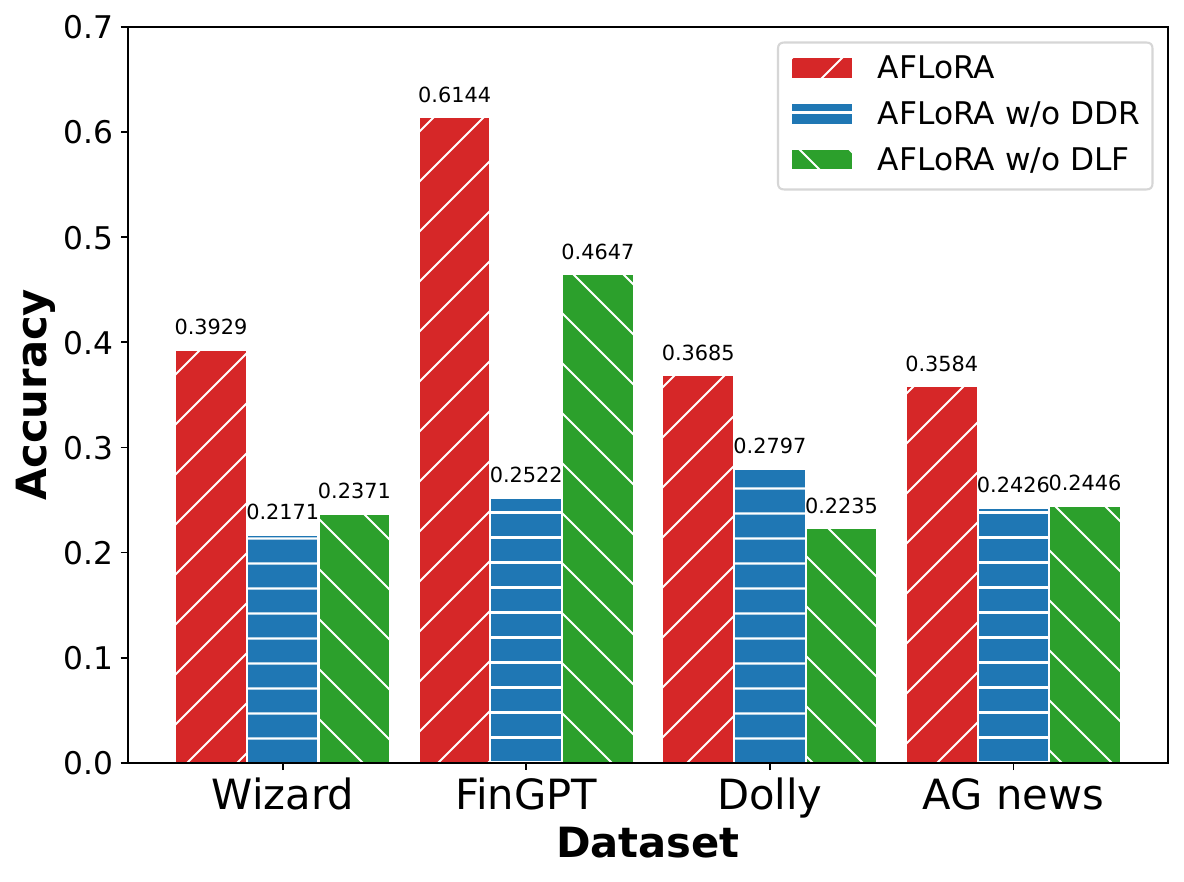}
}
\subfigure[TinyLlama-1.1B]{\label{fig:ablation_llama}
\includegraphics[width=0.625\columnwidth]{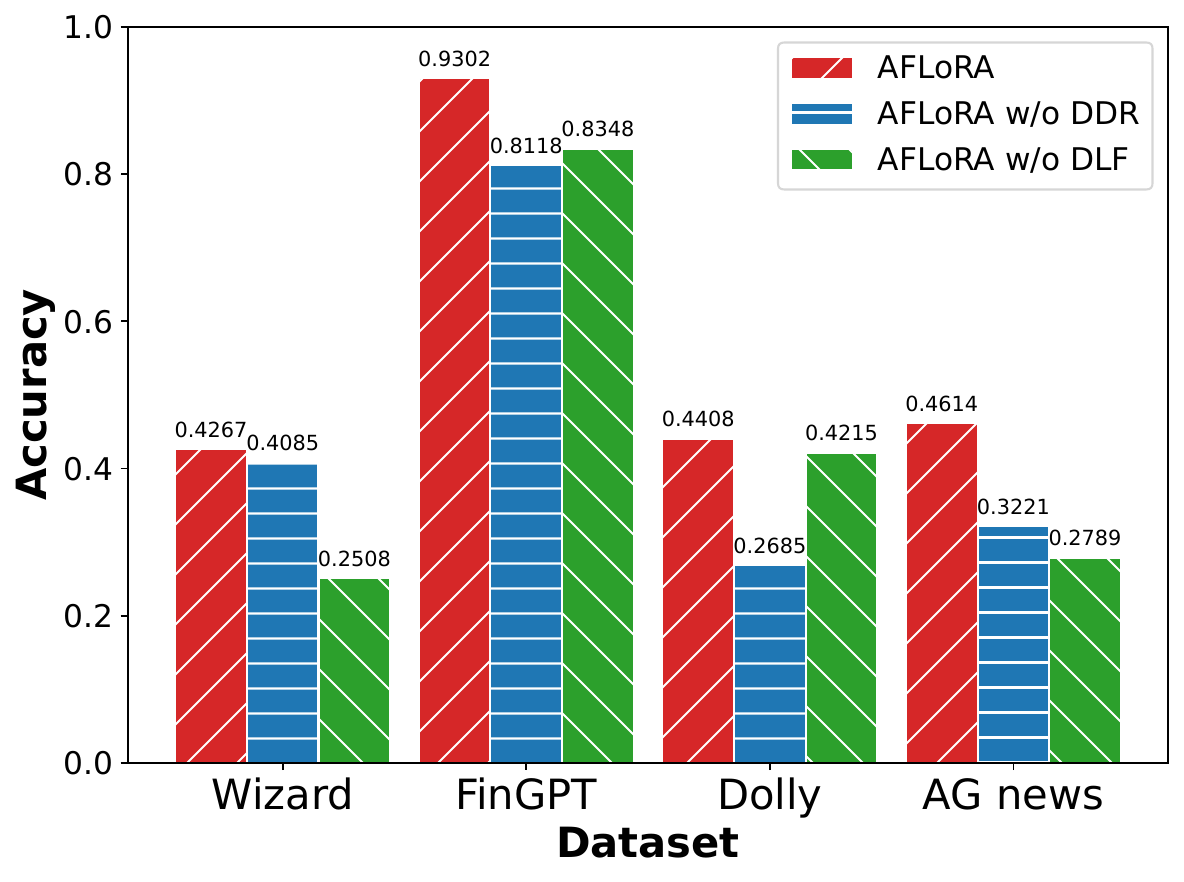}
}
\subfigure[Qwen2.5-3B]{\label{fig:ablation_qwen}
\includegraphics[width=0.625\columnwidth]{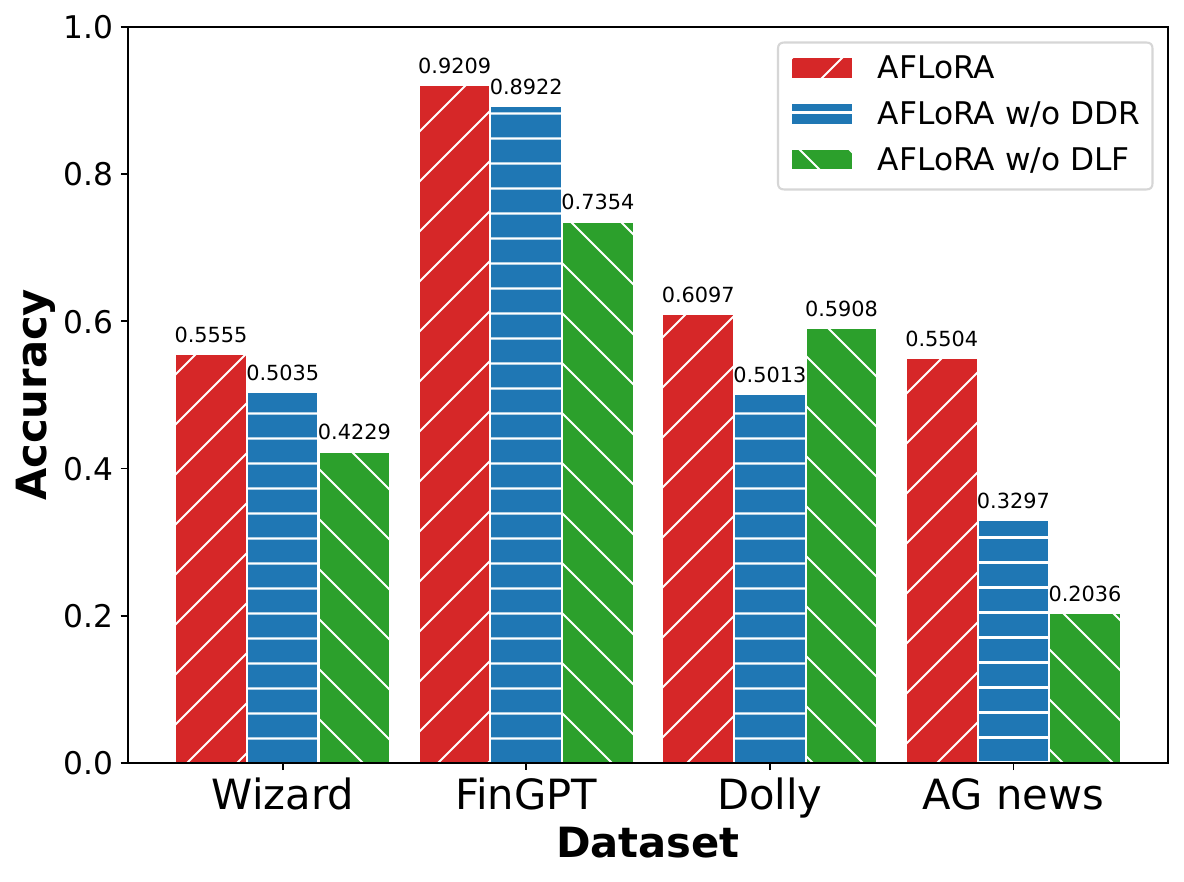}
}
\vspace{-2mm}
\caption{Effect of the proposed mechanisms.}
\label{fig:Ablation}
\vspace{-2mm}
\end{figure*}

\begin{table*}[!t]
\centering
\caption{Performance and cost comparison of AFLoRA and baseline methods under IID scenarios.}
\renewcommand{\arraystretch}{1.3}
\begin{tabular}{c|c|cccc|cccc}
\toprule
\multirow{2}{*}{\textbf{\makecell{Foundation \\ Model}}} & \multirow{2}{*}{\textbf{\makecell{Dataset}}} & \multicolumn{4}{c|}{\textbf{Model Performance}} & \multicolumn{4}{c}{\textbf{Client-Side Costs (\%)}}\\
\cline{3-10}
 & & \textbf{AFLoRA} & \textbf{FLoRA} & \textbf{FlexLoRA} & \textbf{HETLoRA} & \textbf{AFLoRA} & \textbf{FLoRA} & \textbf{FlexLoRA} & \textbf{HETLoRA} \\
\midrule
\multirow{2}{*}{\textbf{GPT-2}} 
& \textbf{Wizard}           &\textbf{0.3929}    &0.3752  &0.2523 &0.1967 &\textbf{0.8095} &1.9027 &1.9027 &1.9027 \\
& \textbf{FinGPT}   &\textbf{0.6144}    &0.5746  &0.2523 &0.3975 &\textbf{0.8062} &1.9027 &1.9027 &1.9027  \\
\midrule
\multirow{2}{*}{\textbf{TinyLlama-1.1B}} 
& \textbf{Wizard}           &\textbf{0.4267}    &0.3882  &0.3319  &0.3636 &\textbf{0.0738} &0.2048 &0.2048 &0.2048 \\
& \textbf{FinGPT}   &\textbf{0.9302}    &0.9284  &0.9218 &0.8246 &\textbf{0.0406} &0.2048 &0.2048 &0.2048  \\
\midrule
\multirow{2}{*}{\textbf{Qwen2.5-3B}} 
& \textbf{Wizard}           &\textbf{0.5555}    &0.4781  &0.5506 &0.5489  &\textbf{0.0349} &0.1229 &0.1229 &0.1229 \\
& \textbf{FinGPT}   &\textbf{0.9209}    &0.9143  &0.9129 &0.8622 &\textbf{0.0289} &0.1229 &0.1229 &0.1229  \\
\bottomrule
\end{tabular}
\label{Table:iid}
\end{table*}

\begin{table*}[!t]
\centering
\caption{Performance and cost comparison of AFLoRA and baseline methods under Non-IID scenarios.}
\renewcommand{\arraystretch}{1.3}
\begin{tabular}{c|c|cccc|cccc}
\toprule
\multirow{2}{*}{\textbf{\makecell{Foundation \\ Model}}} & \multirow{2}{*}{\textbf{\makecell{Dataset}}} & \multicolumn{4}{c|}{\textbf{Model Performance}} & \multicolumn{4}{c}{\textbf{Client-Side Costs (\%)}}\\
\cline{3-10}
 & & \textbf{AFLoRA} & \textbf{FLoRA} & \textbf{FlexLoRA} & \textbf{HETLoRA} & \textbf{AFLoRA} & \textbf{FLoRA} & \textbf{FlexLoRA} & \textbf{HETLoRA} \\
\midrule
\multirow{2}{*}{\textbf{GPT-2}} 
& \textbf{Dolly}   &\textbf{0.3685}    &0.3099  &0.2549 &0.2005 &\textbf{0.6670} &1.9027 &1.9027 &1.9027\\
& \textbf{AG news}            & \textbf{0.3584}       & 0.2900          & 0.2653 &0.2020 &\textbf{0.8918} &1.9027 &1.9027 &1.9027  \\
\midrule
\multirow{2}{*}{\textbf{TinyLlama-1.1B}} 
& \textbf{Dolly}   &\textbf{0.4408}    &0.2878  &0.3211 &0.2840 &\textbf{0.0619} &0.2048 &0.2048 &0.2048\\
& \textbf{AG news}            & \textbf{0.4614}       &0.2508           & 0.3180      & 0.3012 &\textbf{0.0738} &0.2048 &0.2048 &0.2048  \\
\midrule
\multirow{2}{*}{\textbf{Qwen2.5-3B}} 
& \textbf{Dolly}   &\textbf{0.6097}    &0.5626  &0.6015 &0.5664 &\textbf{0.0403} &0.1229 &0.1229 &0.1229\\
& \textbf{AG news}            & \textbf{0.5504}       &0.4728           & 0.2500 &0.2832 &\textbf{0.0443} &0.1229 &0.1229 &0.1229  \\
\bottomrule
\end{tabular}
\label{Table:noniid}
\end{table*}

In this section, we evaluate the performance and efficiency of AFLoRA to validate its effectiveness.

\subsection{Setup}

\textbf{Dataset.} 
We conduct experiments on four datasets widely used in federated fine-tuning experiments.\textbf{WizardLM} \cite{wizard} is a conversational dataset comprising approximately 250,000 instruction-response pairs.
\textbf{FinGPT-Sentiment} \cite{fingpt} is a dataset for financial sentiment analysis, containing over 10,000 annotated entries collected from financial news and social media sources. 
\textbf{Databricks-Dolly-15k} \cite{dolly} is a instruction-following dataset, consisting of 15,000 samples across diverse tasks including open-domain QA, classification, and summarization. 
\textbf{AG News} \cite{agnews} is a benchmark for news classification, containing over 1.2 million articles categorized into four labels.
\textbf{Alpaca} \cite{alpaca} is an instruction-following dataset, comprising 52,000 examples spanning a wide range of user prompts and corresponding responses. We randomly select 2\% of Alpaca to form a small public dataset.

\textbf{Model.} We conduct experiments on four foundation models. \textbf{GPT-2}\cite{gpt} (with 124M parameters) with LoRA applied to a selected set of modules, specifically c\_attn, c\_proj, and c\_fc, as specified in the lora\_target\_modules configuration. \textbf{TinyLlama-1.1B}\cite{tinyllama} and \textbf{Qwen2.5-3B}\cite{qwen} with lora\_target\_modules setting as q\_proj and v\_proj. 

\textbf{FL system and data partition.} Our experiments involve 1 server and 10 clients. Each client is assigned a resource constraint that limits the maximum rank of the LoRA matrices it can fine-tune, set respectively as [64,32,16,16,8,8,4,4,4,4]. When testing performance under different Non-IID degrees, 20\% of the clients are selected to participate in each global round, while in all other experiments, this ratio is set to 100\%.

For the WizardLM and FinGPT-Sentiment datasets, we use IID partitioning, where data is randomly and evenly distributed across all 10 clients. For the Databricks-Dolly-15k and AG News datasets, we adopt Non-IID partitioning: Databricks-Dolly-15k is allocated to clients following the Non-IID strategy of FedIT\cite{fedit} with smaller $\epsilon$ indicating higher heterogeneity. For performance comparison under Non-IID scenarios, we set $\epsilon=0.5$ while for performance comparison under different degrees of Non-IID, we set $\epsilon$ to 0.2, 0.4, and 0.6, respectively. And AG News is split using a label-skew method so that each client receives data with only 2 labels.

\textbf{Metrics.} We evaluate using two metrics: Model Performance and Client-Side Costs. 
Model performance is measured by the accuracy of the global model. For the QA tasks on the WizardLM and Databricks-Dolly-15k datasets, we report accuracy on the MMLU benchmark\cite{mmlu}, a widely used multi-task evaluation suite that covers questions from a range of academic and professional domains. For the financial sentiment analysis task on the FinGPT-Sentiment dataset, model performance is assessed using classification accuracy on the FPB benchmark\cite{fpb}. For the news classification task on the AG News dataset, performance is evaluated on the AG News test set\cite{agnews}. 
Client-side costs are measured as the ratio between the number of parameters locally fine-tuned or communicated from clients to the server, and the total number of parameters in the original LLM. A lower ratio indicates reduced local overhead.

\textbf{Baseline.}
We compare our method with the following representative baselines for LoRA-based federated finetuning:

\begin{itemize}
    \item \textbf{FLoRA}\cite{flora} addresses the problem of accurate aggregation of heterogeneous LoRA in FL. It achieves this by first concatenating the \( \mathbf{B} \) and \( \mathbf{A} \) matrices from all clients and then multiplying the concatenated \( \mathbf{B} \) and \( \mathbf{A} \) to obtain the global update.

    \item \textbf{FlexLoRA}\cite{flexlora} focuses on accurate aggregation of heterogeneous LoRA. Each client $k$ first reconstructs its matrix as \( \mathbf{W}_k = \mathbf{B}_k \mathbf{A}_k \), and weighted model averaging is performed to aggregate the reconstructed matrix across clients. The server then applies SVD on the aggregated \( \mathbf{W} \) to obtain a low-rank approximation that matches each client's specified rank.
    
    \item \textbf{HETLoRA}\cite{hetlora} aims to enable dynamic rank pruning for heterogeneous clients. During local fine-tuning, HETLoRA introduces a penalty term in the loss function to achieve rank self-pruning. To facilitate aggregation across clients with different ranks, it performs zero-padding on the client-specific \( \mathbf{B} \) and \( \mathbf{A} \) matrices to match the maximum rank observed across clients. The zero-padded matrices are then aggregated and multiplied to obtain the global model update.
\end{itemize}

\subsection{Evaluation Results}
\textbf{Effect of the proposed mechanisms.}
To verify the effectiveness of the proposed Diagonal-Based Dynamic Rank Assignment mechanism (DDR) and Decoupled LoRA Fine-Tuning (DLF) mechanism,  we conduct experiments of AFLoRA with and without DDA and DLF, respectively. 
Among them, AFLoRA without DDR initially assigns ranks to each client based on their computation and communication capabilities and maintains these rank assignments throughout the entire fine-tuning process.
For DLF, as prior study\cite{ffalora} has established the redundancy of LoRA fine-tuning, indicating that fine-tuning only the $\mathbf{B}$ matrices yields a well-performing global model, our primary objective is to verify the effectiveness of Server-Side Fine-Tuning. AFLoRA without DLF fine-tunes only the $\mathbf{B}$ at the client side, with no $\mathbf{A}$ fine-tuning performed on the server.
The performance comparison results are shown in Fig.\ref{fig:Ablation}.

It is shown that not performing the DDR mechanism results in model accuracy dropping nearly 15\% on GPT-2, 10\% on TinyLlama-1.1B and Qwen2.5-3B on average. 
This is because the DDR mechanism dynamically adjusts client rankings based on the knowledge acquisition patterns of the LoRA-adapted foundation model on local datasets while simultaneously focusing optimization on established effective directions, thereby preventing overfitting of client-specific LoRA adaptations. 
Furthermore, it selectively retains only the most information-rich dimensions of the LoRA updates, thereby reducing the number of parameters that need to be fine-tuned and uploaded, and consequently lowering client-side overhead.
In conclusion, DDR is an indispensable component for AFLoRA. 

We can also observe that DLF enhances the accuracy of the global model on different foundation models and datasets.
For example, adopting DLF increases the model accuracy by over 12\% on gpt-2 and TinyLlama-1.1B, and almost 17\% on Qwen2.5-3B on average. 
This demonstrates the effectiveness of the DLF mechanism in boosting AFLoRA's performance.  
It employs server-side $\mathbf{A}$ fine-tuning assisted by a small public dataset to prevent the aggregated model from overly biasing toward individual client data distributions, ultimately improving the generalization capability of the global model.

\begin{figure*}[!t]
\centering
\label{fig:noniid}
\subfigure[GPT-2]{\label{fig:noniid_gpt}
\includegraphics[width=0.625\columnwidth]{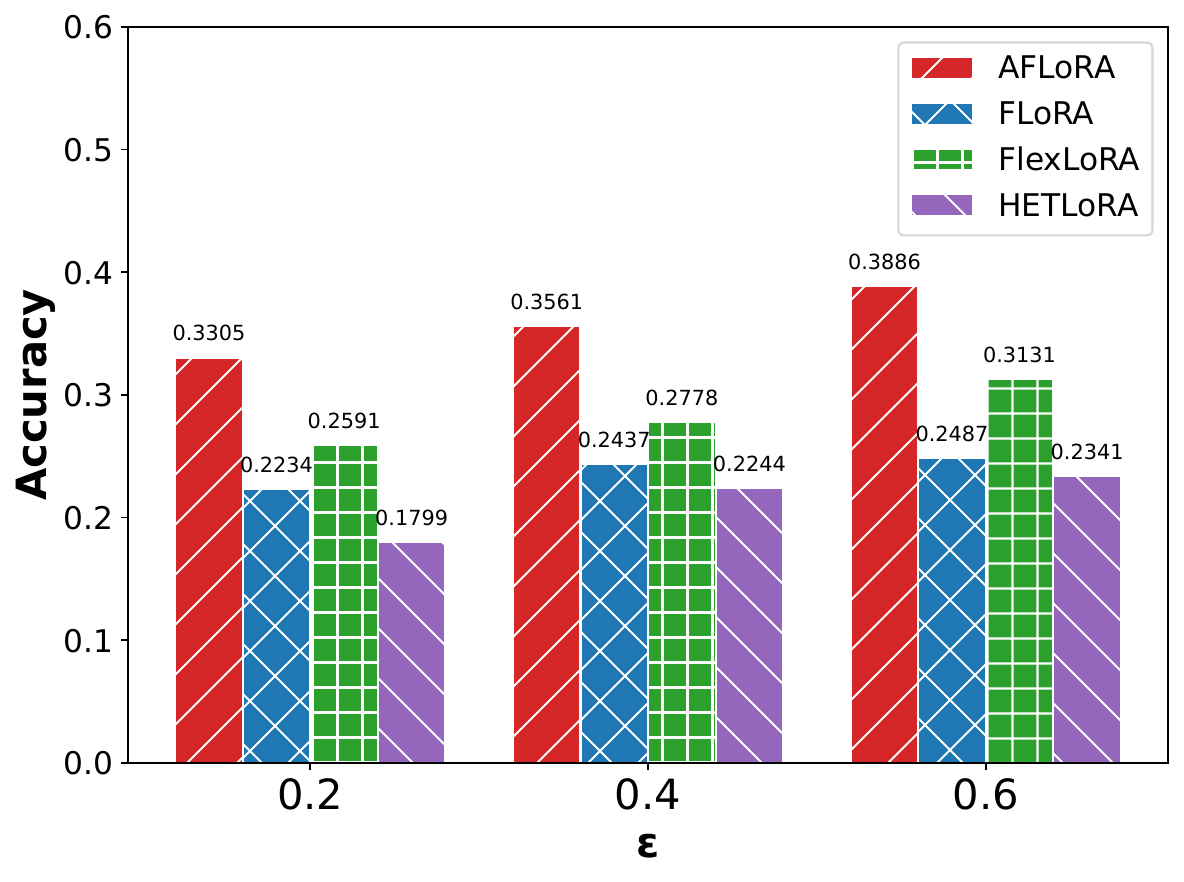}
}
\subfigure[TinyLlama-1.1B]{\label{fig:noniid_llama}
\includegraphics[width=0.625\columnwidth]{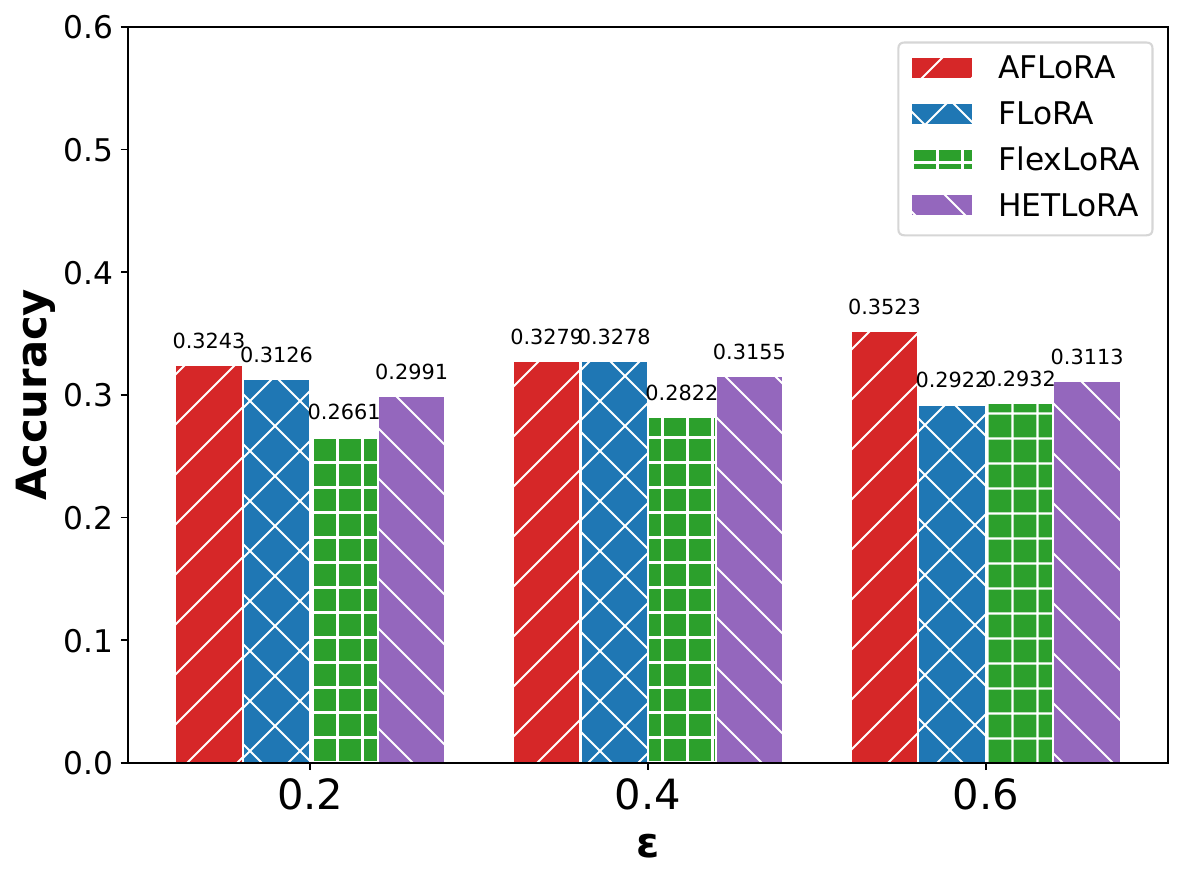}
}
\subfigure[Qwen2.5-3B]{\label{fig:noniid_qwen}
\includegraphics[width=0.625\columnwidth]{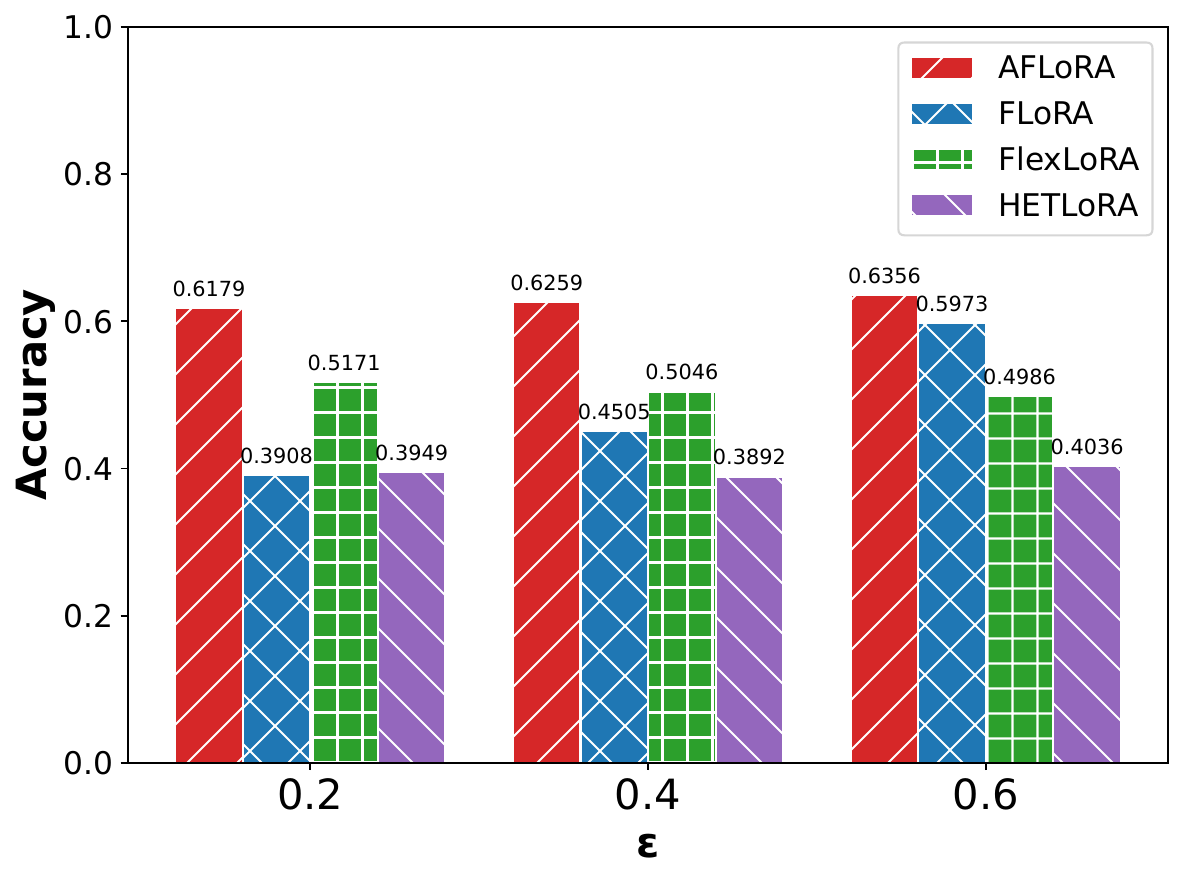}
}
\vspace{-2mm}
\caption{The performance comparison when the degree of non-IID $\epsilon$ changes.}
\label{fig:noniid}
\vspace{-4mm}
\end{figure*}

\textbf{Performance comparison under IID scenarios.}
We compare AFLoRA with baseline methods on the WizardLM and FinGPT-Sentiment datasets under the IID setting, focusing on both global model performance and client-side costs. As shown in Table~\ref{Table:iid}, AFLoRA significantly outperforms existing methods, demonstrating its effectiveness in environments with uniform data and task distributions. 
AFLoRA achieves better global model performance by freezing the shared $\mathbf{A}$ matrix on each client, which stabilizes local updates, preserves shared knowledge, and reduces variance from local adaptation. Fine-tuning the shared $\mathbf{A}$ matrix on the server further helps extract common features, leading to faster convergence and higher final accuracy.
Regarding client-side computation and communication costs, AFLoRA updates and transmits only the client-specific matrices $\mathbf{B}$ and $\mathbf{\Lambda}$, effectively halving resource consumption compared to methods updating both $\mathbf{A}$ and $\mathbf{B}$. The adaptive rank adjustment mechanism further optimizes resource allocation as training progresses. Moreover, since $r \ll \min(m, n)$, the introduced diagonal matrix adds far fewer parameters than the original $\mathbf{A}$ and $\mathbf{B}$.
In contrast, baseline methods exhibit several limitations. All of them do not take the distribution of the data into account. Moreover, FLoRA is unable to adaptively adjust the rank based on the local learning state, resulting in the local model’s inability to effectively adapt to local training. FlexLoRA relies on SVD-based aggregation, introducing additional computational cost and approximation errors that can negatively affect overall performance. HETLoRA does not adequately address interference introduced by aggregation and, due to a large penalty coefficient in the original paper, does not achieve effective rank self-pruning in practice. 
As a result, AFLoRA demonstrates clear advantages in both model quality and efficiency under IID scenarios.

\textbf{Performance comparison under Non-IID scenarios.}
We further evaluate all methods on the Databricks-Dolly-15k and AG News datasets under Non-IID settings, to assess the robustness and generalization of AFLoRA in heterogeneous data settings. As shown in Table~\ref{Table:noniid}, AFLoRA consistently outperforms baseline methods, highlighting its robustness and effectiveness.
Compared to IID scenarios, Non-IID data environments typically lead to a decline in global model performance. This is mainly because the diverse and imbalanced data distributions across clients introduce inconsistencies in local updates, making it more difficult for the aggregated global model to generalize effectively.
AFLoRA maintains strong performance under Non-IID settings for two key reasons. First, the server-side fine-tuning of the shared $\mathbf{A}$ matrix prevents the global representation from overfitting to any specific client’s data distribution, thereby enhancing the model's robustness. Second, AFLoRA’s dynamic rank adjustment mechanism allocates LoRA capacity adaptively based on each client's local data distribution and learning progress, making it better suited for heterogeneous data and allowing each client to effectively contribute according to the complexity of its local data.
Besides the aforementioned shortcomings, FLoRA, FlexLoRA and HETLoRA cannot adaptively adjust the rank in accordance with the local data and learning state, and directly aggregate the local updates which amplifies bias toward clients with dominant distributions, further compromising global model quality in Non-IID conditions.

\textbf{Performance under different non-IID degrees.}
We evaluate our method under varying degrees of data heterogeneity for the Databricks-Dolly-15k dataset, with results presented in Fig.\ref{fig:noniid}.
AFLoRA consistently outperforms competing approaches in model performance regardless of the degree of data heterogeneity, demonstrating remarkable stability, maintaining high accuracy levels across all $\epsilon$ settings. In contrast, baseline methods suffer substantial performance degradation under high data heterogeneity (low $\epsilon$ values), highlighting AFLoRA's robustness in non-IID scenarios.
This can be attributed to the fact that higher data heterogeneity exacerbates the negative impact of uneven learning on the global model's generalization capabilities. While baseline methods struggle under these conditions, AFLoRA effectively mitigates such issues through its server-side fine-tuning and dynamic rank assignment, resulting in significantly better performance, particularly in challenging Non-IID environments.
Additionally, we evaluate the model on the general dataset to assess its performance on standard tasks, measuring the algorithm's overall effectiveness when task relevance is broad and data diversity is high, which is more closely resemble large-scale, real-world applications spanning multiple user tasks.

\section{Conclusion}\label{sec:conclusion}

In this paper, we presented AFLoRA, an adaptive and resource-efficient federated fine-tuning framework designed to tackle the challenges of heterogeneity and limited resources in real-world environments.
We introduced a diagonal-based dynamic rank assignment that enables clients to adjust LoRA ranks in response to their local resources and adaptation states, guided by a learnable diagonal matrix. 
By decoupling LoRA matrix updates between clients and the server, we effectively reduced client-side communication and computation overhead. Furthermore, we proposed a zero-padding-based rank-aware aggregation mechanism to ensure accurate and lossless integration of heterogeneous client updates. 
Extensive experiments demonstrated that AFLoRA achieves superior adaptation performance and fine-tuning efficiency compared to state-of-the-art methods, making it a practical and effective solution for federated LLM adaptation in heterogeneous and resource-constrained environments.

\bibliographystyle{IEEEtran}
\bibliography{reference}
\end{document}